\pdfoutput=1

\documentclass[11pt]{article}

\usepackage[preprint]{acl}

\usepackage{times}
\usepackage{latexsym}
\usepackage{colortbl} 
\usepackage{multirow} 
\usepackage{xcolor}  
\usepackage{booktabs} 
\usepackage{amsmath}
\definecolor{lightpink}{rgb}{0.945, 0.816, 0.804}
\definecolor{lightgreen}{rgb}{0.851, 0.906, 0.839}
\definecolor{lightblue}{rgb}{0.8, 0.9, 1}
\definecolor{lightyellow}{rgb}{0.992, 0.949, 0.816}
\usepackage[T1]{fontenc}

\usepackage[utf8]{inputenc}

\usepackage{microtype}
\usepackage{float}
\usepackage{inconsolata}

\usepackage{graphicx}
\usepackage{authblk}
\newcommand{\shortname}{LLaVA-Med}
\title{OmniV-Med: Scaling Medical Vision-Language Model\\ for Universal Visual Understanding}

\author{
  Songtao Jiang$^{1}$, Yuan Wang$^{1}$, Sibo Song$^{2}$, Yan Zhang$^{1}$,
  Zijie Meng$^{1}$,\\
   Bohan Lei$^{1}$, Jian Wu$^{1}$,
  Jimeng Sun$^{3}$, Zuozhu Liu$^{1\dagger}$\\
  $^1$Zhejiang University \quad
  $^2$Alibaba Group \quad
  $^3$UIUC
}

\begin{document}
\maketitle
\let\thefootnote\relax\footnote{† Corresponding author: zuozhuliu@intl.zju.edu.cn}

\begin{abstract}

The practical deployment of medical vision-language models (Med-VLMs) necessitates seamless integration of textual data with diverse visual modalities, including 2D/3D images and videos, yet existing models typically employ separate encoders for different modalities. To address this limitation, we present OmniV-Med, a unified framework for multimodal medical understanding. Our technical contributions are threefold: First, we construct OmniV-Med-Instruct, a comprehensive multimodal medical dataset containing 252K instructional samples spanning 14 medical image modalities and 11 clinical tasks. Second, we devise a rotary position-adaptive encoder that processes multi-resolution 2D/3D images and videos within a unified architecture, diverging from conventional modality-specific encoders. Third, we introduce a medical-aware token pruning mechanism that exploits spatial-temporal redundancy in volumetric data (e.g., consecutive CT slices) and medical videos, effectively reducing 60\% of visual tokens without performance degradation. Empirical evaluations demonstrate that OmniV-Med-7B achieves state-of-the-art performance on 7 benchmarks spanning 2D/3D medical imaging and video understanding tasks. Notably, our lightweight variant (OmniV-Med-1.5B) attains comparable performance while requiring only 8 RTX3090 GPUs for training and supporting efficient long-video inference. Data, code and model will be released.
\end{abstract}

\section{Introduction}
\begin{figure}[!t]
    \centering
    \includegraphics[width=0.9\linewidth]{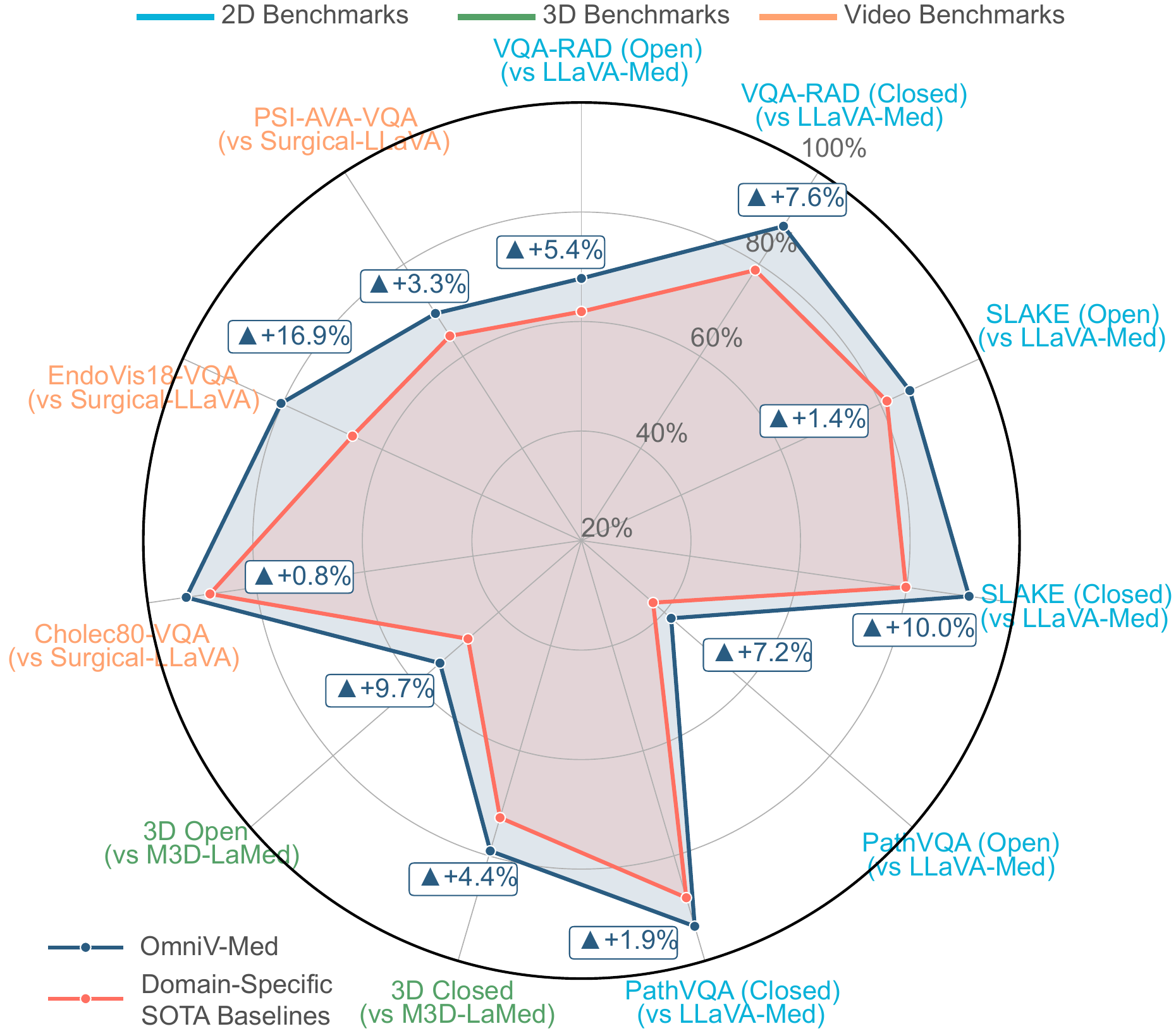}
\caption{Comparison with state-of-the-art methods on medical 2D/3D image and video benchmarks.}
    \label{fig:intro}
\end{figure}
The unified understanding and generation of textual and visual inputs is pivotal for advancing next-generation vision-language models (VLMs)~\citep{bordes2024introductionvisionlanguagemodeling,bai2023qwen,wu2024next,zhan2024anygpt,zhang2025videollama}. Recent progress in multimodal VLMs has demonstrated promising results, with frameworks leveraging unified tokenizers or specialized encoders to process heterogeneous inputs (e.g., images, text, audio, and video)~\citep{Wu2023NExTGPTAM,Zhan2024AnyGPTUM,li2024llavanextinterleavetacklingmultiimagevideo}. In the medical domain, emerging works such as LLaVA-Med, RadFM, and PathChat~\citep{li2024llava,wu2023towards,Lu2024AMG,Xu2024AWF} have extended these capabilities to medical vision-language tasks. However, existing approaches either focus on domain-specific image interpretation~\citep{Zhou2023AFM,Jiao2023USFMAU,Jiang2024MedMoEMO,Moor2023MedFlamingoAM,li2024llava}, or are restricted to radiology-specific 2D/3D imaging analysis~\citep{wu2023towards}. This fragmentation limits their ability to holistically process multimodal visual data with textual instructions (e.g., integrating endoscopic video, 3D CT scans and 2D images), thereby hindering deployment in complex real-world scenarios such as AI-assisted surgery and adaptive treatment planning~\citep{bodenstedt2020artificial,wang2023foundation,zhang2024generalist}.

Developing a unified Med-VLM model capable of processing diverse multimodal inputs poses three fundamental challenges. First, the medical domain lacks robust encoders tailored for 3D volumetric data and temporal videos~\citep{wang2022medical,nwoye2023cholectriplet2021}, compounded by the absence of unified encoders capable of simultaneously modeling both modalities. Discrepancies among 2D/3D spatial structures and temporal video dynamics, along with limited high-quality medical datasets, pose significant challenges in achieving robust alignment of cross-modal representation~\citep{he2024unified,ren2024medical}. Second, widely-used encoders like CLIP-ViT~\citep{radford2021learning}, which rely on fixed positional embeddings, struggle to adapt to the diverse spatial resolutions (e.g., gigapixel pathology slides) and temporal dynamics in medical videos. Additionally, existing tokenization strategies for medical 3D volumes and videos are inefficient and lack scalability~\citep{hirsch2023self}. Although techniques such as bilinear interpolation~\citep{li2024llavaone,kirkland2010bilinear} have been used in general domains to mitigate this issue, they often fail to preserve the fine-grained details essential for medical analysis. Third, the medical domain suffers from a severe shortage of instruction-aligned datasets for 3D and video modalities. The lack of large-scale, task-specific training data hinders the development of models capable of robust multimodal alignment and domain-adaptive reasoning, limiting their practical utility in clinical settings.

In this paper, we propose \textbf{OmniV-Med}, a unified Med-VLM foundation model for processing medical text, 2D/3D images, and medical videos. We first construct \textbf{OmniV-Med-Instruct}, a comprehensive instruction-following dataset spanning 14 medical  (e.g., 2D single/multi-image, 3D volumetric, and video data) with 252K vision-language samples.
For the model architecture, unlike prior methods that rely on modality-specific encoders~\citep{li2023videochat,lin2023video}, we design a single vision encoder leveraging a rotary position-adaptive encoding strategy~\citep{dehghani2023patch} to unify 2D, 3D, and video processing.
Our encoder first learns relative spatial relationships between patches in 2D images (height/width dimensions) and extrapolates these positional encodings to the third dimension: 1) 3D images are treated as multi-view slices with spatial position encoding, and 2) videos are modeled as multi-frame sequences with temporal position encoding. This contrasts with conventional approaches that simply concatenate fixed positional embeddings for multi-image 3D/video inputs, lacking explicit spatiotemporal modeling.
Furthremore, we adopt a progressive strategy: the model is first pretrained on high-quality 2D vision-language pairs, then extended to 3D and video data using spatiotemporal extrapolation. 
This progressive training enables emergent capabilities, facilitating fast and effective cross-modal alignment by projecting all modalities into a unified latent space.

To reduce computational overheads, we introduce a medical-aware token pruning strategy that dynamically compresses redundant tokens during training and inference. By computing $L-1$ distances between patch of consecutive video frames or adjacent 3D slices, we prune tokens with differences below a predefined threshold~\citep{choudhury2024don}. This encourages the model to prioritize significant frame-to-frame and cross-view variations, demonstrating satisfactory trade-off between performance and computational efficiency. 

Our method surpasses state-of-the-art(SoTA) task-specific models across 7 widely-used medical vision-language benchmarks across 2D/3D images and videos. For 2D medical visual question answering (VQA), we evaluate on both close- and open-end tasks in Rad-VQA~\citep{lau2018dataset}, SLAKE~\citep{liu2021slake}, and PathVQA~\citep{he2020pathvqa}. For 3D tasks, we assess performance on the latest M3D-VQA~\citep{bai2024m3d}, covering both close- and open-ended VQA tasks. For medical video, we evaluate on the Cholec80-VQA~\citep{seenivasan2022surgical}, PSI-AVA-VQA~\citep{valderrama2022towards}, and EndoVis18-VQA~\citep{seenivasan2023surgicalgpt}. 
Our OmniV-Med establishes SoTA performance on all 7 benchmarks with \textbf{OmniV-Med-7B}. Our lightweight version, \textbf{OmniV-Med1.5B}, achieves competitive performance while requiring only 8$\times$ RTX 3090 GPUs for training, matching or even surpassing SoTA models in individual tasks. To our knowledge, this work presents the first Med-VLM unifying text, 2D/3D images, and videos within a single framework, offering a scalable and practical solution for real-world clinical workflows.
\begin{figure*}[!ht]
    \centering
    \includegraphics[width=1\linewidth]{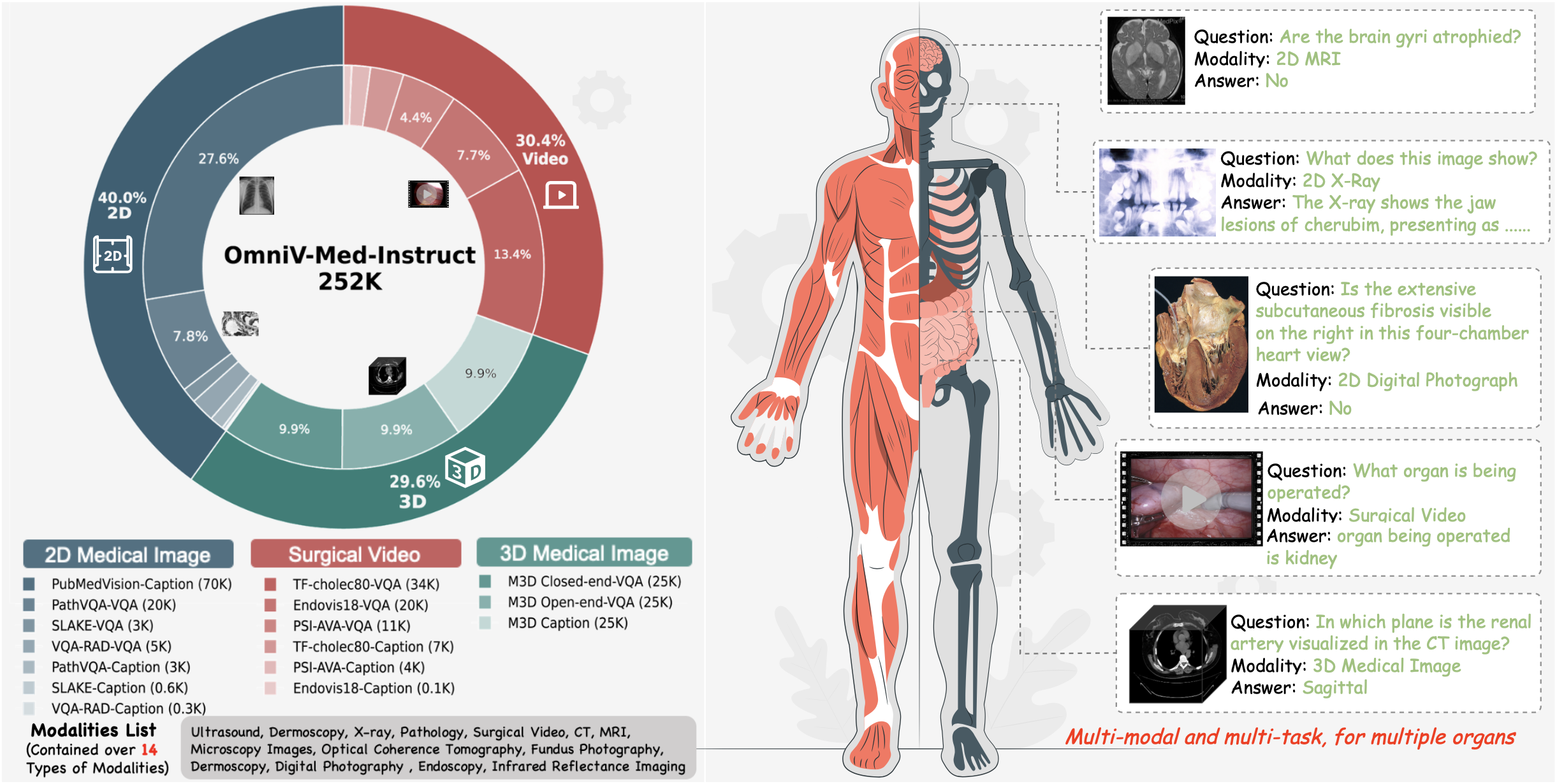}
\caption{Data distribution, coverage and examples of OmniV-Med-Instruct.}
    \label{fig:data}
\end{figure*}
\section{OmniV-Med}
\subsection{OmniV-Med-Instruct Construction}

The coverage of data modalities, sample sizes, tasks and statistics in OmniV-Med-Instruct are illustrated in Figure~\ref{fig:data}. Our dataset covers 14 imaging modalities for various diseases, including regular images such as CT, MRI, X-ray, and pathology, and domain-specific images such as dermoscopy and endoscopy. For 2D-VL instructions, we curated mainstream medical datasets on two major tasks: Medical VQA (Med-VQA) and Medical Report Generation (MRG), including Rad-VQA~\citep{lau2018dataset}, MIMIC-CXR~\citep{bae2024mimic}, SLAKE~\citep{liu2021slake}, and PathVQA~\citep{he2020pathvqa}. To enhance the model's instruction-following capabilities, we also sampled 70K instruction-following examples from PubMedVision~\citep{chen2024huatuogpt}. For 3D-VL pairs, we sampled 75K data points from the M3D datasets~\citep{bai2024m3d}, which cover diverse 3D medical tasks such as image-text retrieval, MRG, VQA, and positioning.

\begin{figure*}[!htbp]
    \centering
    \includegraphics[width=1\linewidth]{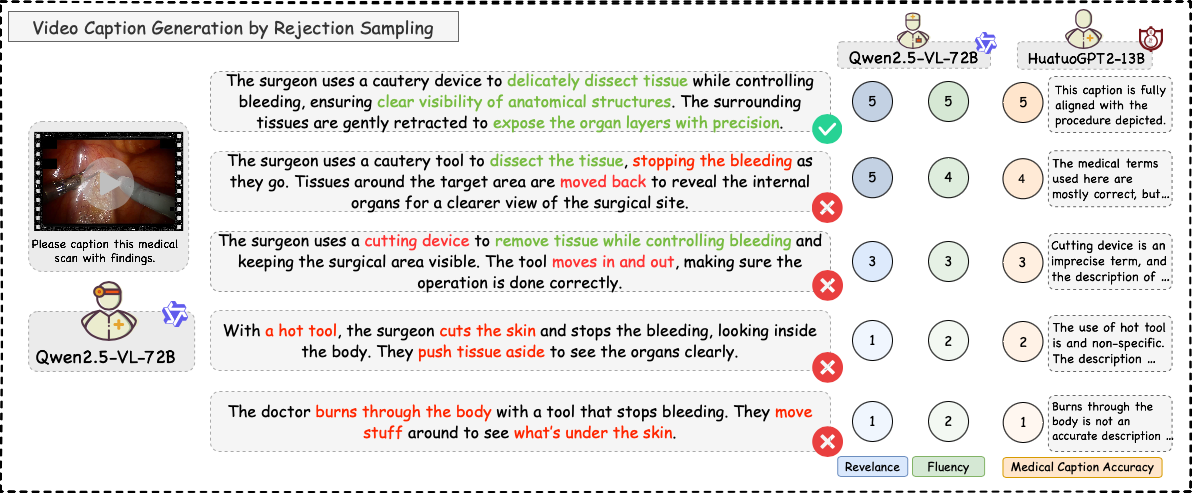}
\caption{A framework for high-quality medical video captioning via rejection sampling: Utilizing Qwen2.5-VL-72B for candidate generation and combining Qwen2.5-VL-72B with HuatuoGPT2-13B for medical caption evaluation based on relevance, fluency, and accuracy (5-point scale).}
    \label{fig:rejction_sampling}
\end{figure*}

For medical video-language pairs, we leveraged existing VLMs to build the an open-source medical video instruction-following dataset. First, we created video VQA tasks using datasets including Cholec80~\citep{seenivasan2022surgical}, PSI-AVA~\citep{valderrama2022towards}, and EndoVis18~\citep{seenivasan2023surgicalgpt}, which cover various surgical procedures. We formed short video segments from consecutive frames and their original VQA questions and answers, allowing the model to learn from continuous video sequences. For video captioning tasks, due to the absence of open-source medical-specific video VLMs, we used the general Qwen2.5-VL-72B~\citep{bai2025qwen2} to generate captions by combining frame-level QA texts with their corresponding video segments. To ensure data quality, we applied rejection sampling~\citep{touvron2023llama} to filter out high-quality video-text pairs (see details in Figure~\ref{fig:rejction_sampling}).


\begin{figure*}[!t]
    \centering
    \includegraphics[width=1\linewidth]{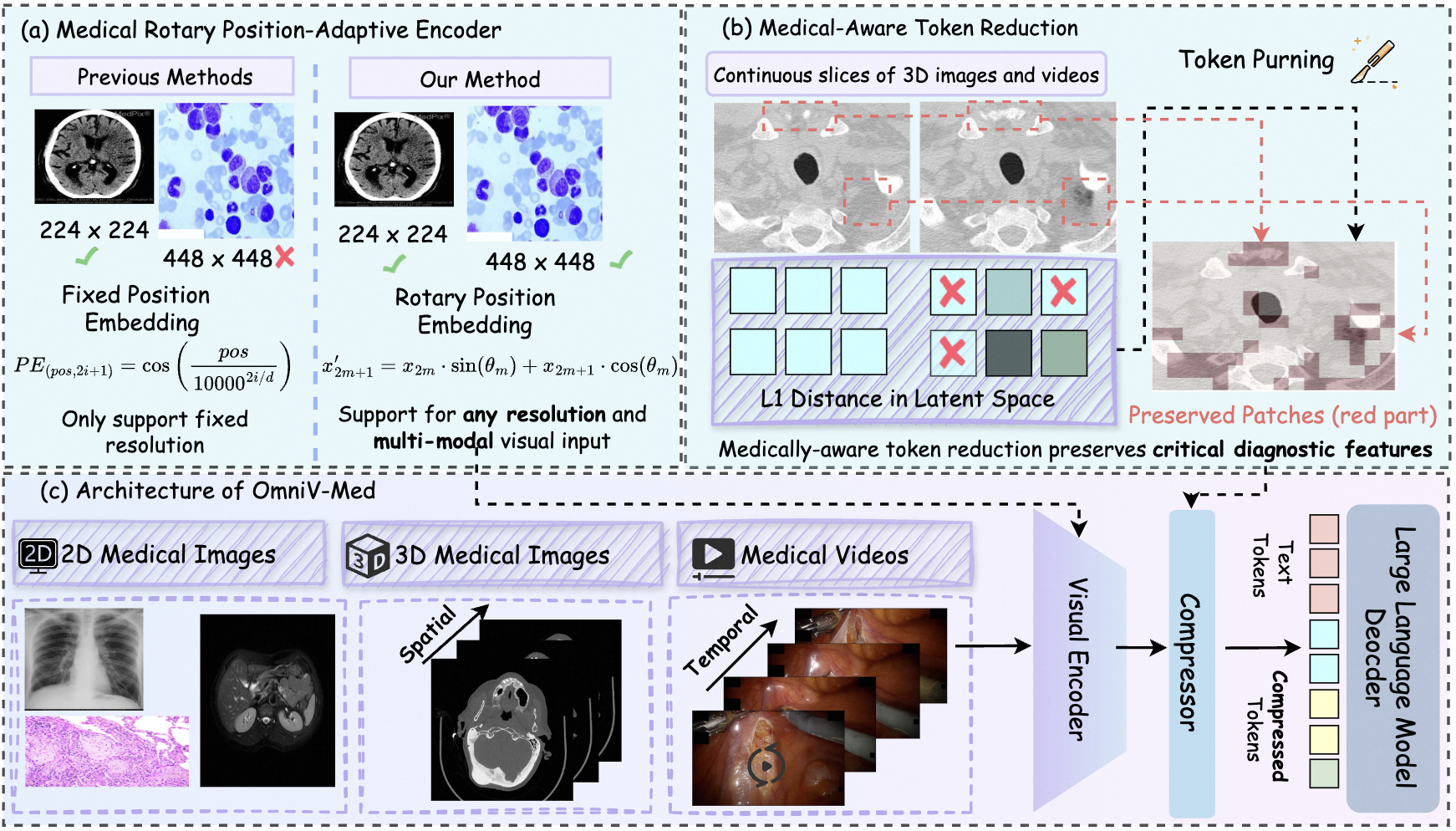}
\caption{OmniV-Med framework consists of three key components: (a). a medical rotary position-adaptive encoder supporting various multi- modalities with different resolutions, (b). medical-aware token reduction to efficiently handle redundant frames and slices in videos and 3D images, and (c). the architecture of our OmniV-Med model.}
    \label{fig:pipeline}
\end{figure*}
\subsection{Training Pipeline}

Our training process consists of three stages: 1) 2D Med-VL alignment, enhancing the model's understanding of 2D medical images with VL caption pairs; 2) medical instruction-following tuning, enabling the model to follow complex instructions based on medical image understanding; and 3) medical mixed-modal tuning, improving model’s capacity to interpret diverse spatial and temporal VL inputs.


\noindent\textbf{2D Med-VL Alignment.}
In stage 1, we align medical vision and language inputs using 647K image-caption pairs from \texttt{PubMedVision\_Alignment}~\citep{chen2024huatuogpt}. We freeze the LLM backbone and fine-tune the pre-trained \texttt{SigLIP}~\citep{zhai2023sigmoid} along with the MLP projector as the vision encoder. To better handle different image resolutions, we replace the original fixed position embedding in \texttt{SigLIP} with a rotary position embedding, creating the \textit{rotary position-adaptive encoder}.  The rotary position embedding is defined as: 
\begin{equation}
\mathbf{R}_{\theta, d} = \begin{pmatrix}
\cos \theta & -\sin \theta \\
\sin \theta & \cos \theta
\end{pmatrix},
\end{equation}
where $\theta$ represents the rotation angle and $d$ denotes the embedding dimension. For a given position $m$, the rotary embedding is applied to the query $\mathbf{q}_m$ and key $\mathbf{k}_n$ vectors as:
\begin{equation}
\mathbf{q}_m' = \mathbf{R}_{\theta, d}(m) \mathbf{q}_m, \quad \mathbf{k}_n' = \mathbf{R}_{\theta, d}(n) \mathbf{k}_n.
\end{equation}
This approach enhances the model's ability to capture relative positional relationships, improving its adaptability to diverse medical visual inputs.

\noindent\textbf{Medical Instruction-Following Tuning.}  
In stage 2, we make all model parameters trainable to improve the model’s capability in following complex textual instructions based on visual inputs. We train the model using 647K instruction-answer pairs from \texttt{PubMedVision\_SFT}~\citep{chen2024huatuogpt}. Both stage 1 and stage 2 focus on 2D images, as our experiments indicate that a strong foundational understanding of 2D images is crucial to extend Med-VLMs to complex 3D images and videos. By the end of this stage, the model can effectively handle various visual tasks related to 2D images.

\noindent\textbf{Medical Mixed-Modal Tuning.}  
In stage 3, the goal is to expand the model's capabilities from 2D images to 3D images (spatial) and videos (temporal). However, processing 3D images and videos is computationally expensive. In response, we introduce a token pruning mechanism to remove redundant information while preserving critical content. We found that consecutive slices in 3D medical images and adjacent frames in medical videos often contain repetitive or similar information. Hence, we compressed visual tokens by calculating the $L1$ distance between patches of consecutive views or frames. Tokens with an $L1$ distance below a threshold of $0.1$ are identified as redundant and removed. This pruning method effectively reduces unnecessary visual data without impacting model performance. 
The model is trained with all 252K samples in \texttt{OmniV-Med-Instruct} with a standard autoregressive loss. All parameters are trainable to fully unleash the model's potential, enabling it to extrapolate visual understanding to spatiotemporal dimensions and enhance downstream performance across diverse tasks. Our approach achieves SoTA accuracy with improved efficiency, enhancing its practicality for real-world medical tasks.

\begin{table*}[ht!]
\centering
\renewcommand{\arraystretch}{1}
\resizebox{1\textwidth}{!}{
\begin{tabular}{ll|cc|cc|cc|c}  
 & & \multicolumn{2}{c|}{ VQA-RAD} & \multicolumn{2}{c|}{ SLAKE} & \multicolumn{2}{c|}{ PathVQA}  \\
Method  &  & Open   & Closed    & Open   & Closed  & Open &  Closed & \\
\hline
\multicolumn{9}{l}{\it Representative \& SoTA methods with numbers reported in the literature (Non-VLM Based Methods) } \\  
\hline
\rowcolor[HTML]{F5F5F5}VL Encoder–Decoder~\citep{bazi2023vision} & & - & 82.47 & - & - & - & 85.61  \\
\rowcolor[HTML]{F5F5F5}Q2ATransformer~\citep{liu2023q2atransformer} & & - & 81.20 & - & - & 54.85 & 88.85  \\
\rowcolor[HTML]{F5F5F5}Prefix T. Medical LM~\citep{van2023open} & & - & - & - & 82.01 & - & 87.00 \\
\rowcolor[HTML]{F5F5F5}PubMedCLIP~\citep{eslami2023pubmedclip} & & - & 80.00 & - & 82.50 & - & - \\
\rowcolor[HTML]{F5F5F5}BiomedCLIP~\citep{zhang2023large} & & - & 79.80 & - & 89.70 & - & - \\
\rowcolor[HTML]{F5F5F5}M2I2~\citep{li2022self} & & - & 83.50 & - & 91.10 & - & 88.00  \\
\rowcolor[HTML]{F5F5F5}BiomedGPT-S~\citep{zhang2023biomedgpt} & & 13.40 & 57.80  & 66.50 & 73.30 & 10.70 & 84.20\\
\rowcolor[HTML]{F5F5F5}BiomedGPT-M~\citep{zhang2023biomedgpt}& & 53.60 & 65.07 & 78.30 & 86.80 & 12.5 & 85.70  \\
\rowcolor[HTML]{F5F5F5}CLIP-ViT w/ GPT2-XL& & - & - & 84.30 & 82.10 & 40.0 & 87.00 \\
\hline
\multicolumn{9}{l}{\it Results (VLM Based Methods)} \\
\hline
\rowcolor[HTML]{F5F5F5}LLaVA~\citep{Liu_2024_CVPR} & & 50.00 & 65.07 & 78.18 & 63.22 & 7.74 & 63.20 \\
\rowcolor[HTML]{F5F5F5}Med-Flamingo~\citep{moor2023med} & & 50.00 & 65.07 & 78.18 & 63.22 & 7.74 & 63.20 \\
\rowcolor[HTML]{F5F5F5}HuatuoGPT-Vision-34B~\citep{chen2024huatuogpt} & & - & 63.80 & - & 74.50 & - & 59.90 \\
\rowcolor[HTML]{F5F5F5}\shortname{} (LLama7B)~\citep{li2024llava} & & {61.52} & { 84.19} & 83.08 & {85.34}  &  {37.95}   & {91.21}  \\
\rowcolor[HTML]{F5F5F5}\shortname{} (Vicuna7B)~\citep{li2024llava} & &  \underline{64.39} & 81.98 & {84.71} & 83.17  &   38.87  & {91.65} \\
\rowcolor[HTML]{F5F5F5}Med-MoE (Phi2-3.6B)~\citep{jiang2024med} & & {58.55} & {82.72} &   \underline{85.06} &  85.58 & 34.74 &  91.98  \\
\rowcolor[HTML]{F5F5F5}Med-MoE (StableLM-2B)~\citep{jiang2024med} & & 50.08 & 80.07 &  {83.16} & {83.41} &  33.79 & 91.30  \\
\rowcolor[HTML]{e3f2fd}OmniV-Med-Tiny (Qwen2.5-1.5B) & & {59.22} & \underline{84.93} &  {84.59} & \underline{87.98} &  \underline{38.87}& \underline{92.66}  \\
\rowcolor[HTML]{e3f2fd}OmniV-Med (Qwen2.5-7B) & & \textbf{67.87} &  \textbf{88.24} &  \textbf{85.92} & \textbf{91.51} &  \textbf{41.68} &  \textbf{93.42} \\
\hline
\end{tabular}
}
\caption{Performance on Med-VQA tasks. \textbf{Bold} denotes the best performance; \underline{underlined} denotes the second-best.}
\label{tab:main_table}
\end{table*}

\begin{figure*}[!t]
    \centering
    \includegraphics[width=1\linewidth]{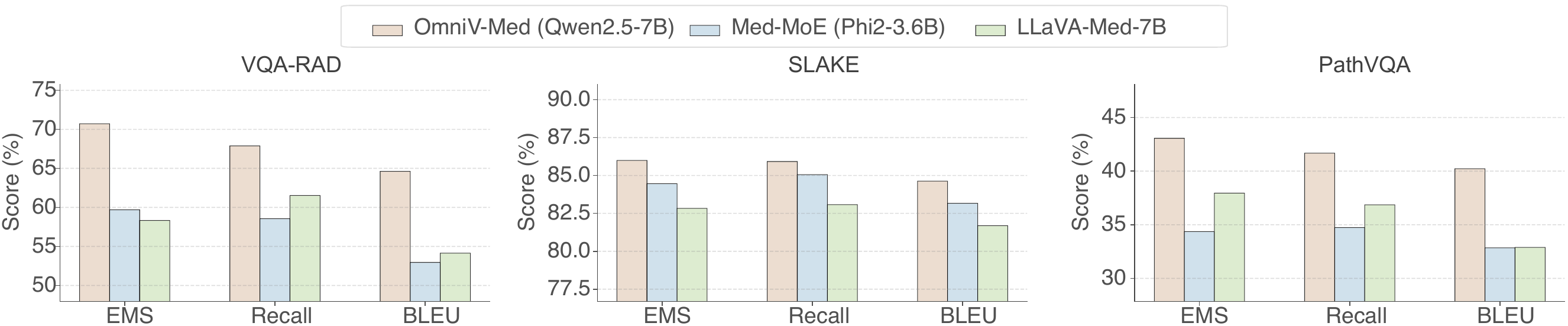}
\caption{Detailed comparison with additional metrics in open settings. \textbf{EMS} denotes \textit{Exact Match Score}.}
    \label{fig:performance on open-end}
\end{figure*}

\begin{table*}[!t]
    \centering
    \small
    \renewcommand{\arraystretch}{1}
    \begin{tabular}{l|ccccc|c}
        \toprule
        \textbf{Methods} & \textbf{Plane} & \textbf{Phase} & \textbf{Organ} & \textbf{Abnormality} & \textbf{Location} & \textbf{Avg.} \\
        \midrule
        \textbf{Closed-End (Accuracy)} \\
        \midrule
       \rowcolor[HTML]{F5F5F5} RadFM~\citep{wu2023towards}  & 19.65 & 28.70 & 16.80 & 18.92 & 14.88 & 19.79 \\
       \rowcolor[HTML]{F5F5F5} M3D-LaMed~\citep{bai2024m3d} & \underline{98.80} & 79.75 & 74.75 & \underline{66.65} & 58.94 & 75.78 \\
       \rowcolor[HTML]{e3f2fd}OmniV-Med-Tiny (Qwen2.5-1.5B) & \textbf{98.95} & \underline{91.35} & \underline{75.15} & 66.24 & \underline{60.78} & \underline{78.49} \\
        \rowcolor[HTML]{e3f2fd}OmniV-Med (Qwen2.5-7B) & {98.75} & \textbf{91.65} & \textbf{76.35} & \textbf{66.80} & \textbf{61.86} &\textbf{79.08} \\
        \midrule
        \textbf{Open-End (Recall)} \\
        \midrule
       \rowcolor[HTML]{F5F5F5}
        RadFM~\citep{wu2023towards} & 14.24 & 14.25 & 14.24 & 15.64 & 23.58 & 16.39 \\
        \rowcolor[HTML]{F5F5F5}
        M3D-LaMed-7B~\citep{bai2024m3d} & \underline{98.37} & 74.41 & 34.20 & 15.91 & 24.00 & 49.38 \\
        \rowcolor[HTML]{e3f2fd}
        OmniV-Med-Tiny (Qwen2.5-1.5B) & 98.18 & \textbf{88.80} & \underline{40.62} & \underline{16.80} & \underline{26.04} & \underline{54.09} \\
        \rowcolor[HTML]{e3f2fd}
        OmniV-Med (Qwen2.5-7B) & \textbf{98.84} & \underline{86.70} & \textbf{41.36} & \textbf{17.26} & \textbf{26.75} & \textbf{54.18} \\
        \bottomrule
    \end{tabular}
    \caption{Performance Comparison of Methods in M3D Tasks.}
    \label{tab:performance_comparison_3d}
\end{table*}

\begin{table*}[ht!]
    \centering
    \renewcommand{\arraystretch}{1}
    \begin{tabular}{l|ccc}
        \toprule
        \textbf{Methods} & \textbf{Cholec80-VQA} & \textbf{EndoVis18-VQA} & \textbf{PSI-AVA-VQA} \\
        \midrule
       \rowcolor[HTML]{F5F5F5} VisualBert~\citep{li2019visualbert}  & 89.7 & 61.4 & 58.5 \\
       \rowcolor[HTML]{F5F5F5} Block~\citep{ben2019block}& 89.5 & 60.1 & 59.9 \\
       \rowcolor[HTML]{F5F5F5} MFH~\citep{yu2018beyond} & 87.5 & 58.8 & 47.8 \\
       \rowcolor[HTML]{F5F5F5}\rowcolor[HTML]{F5F5F5} Surgical-VQA~\citep{seenivasan2022surgical}  & 89.8 & 63.2 & 65.6 \\
       \rowcolor[HTML]{F5F5F5} LV-GPT~\citep{seenivasan2023surgicalgpt} & 87.5 & 68.1 & 59.3 \\
        \rowcolor[HTML]{F5F5F5}Surgical-LLaVA~\citep{jin2024surgicalllavasurgicalscenariounderstanding} & \underline{92.2} & 68.7 & \underline{67.1} \\
        \rowcolor[HTML]{e3f2fd}OmniV-Med-Tiny (Qwen2.5-1.5B)& {91.7} & \underline{78.8} & {64.8} \\
        \rowcolor[HTML]{e3f2fd}OmniV-Med (Qwen2.5-7B)& \textbf{92.9} & \textbf{80.3} & \textbf{69.3} \\
        \bottomrule
    \end{tabular}
     \caption{Performance Comparison of Methods on Different Video Benchmarks.}
    \label{tab:performance_comparison_video}
\end{table*}
\begin{figure}[!t]
    \centering
    \includegraphics[width=1\linewidth]{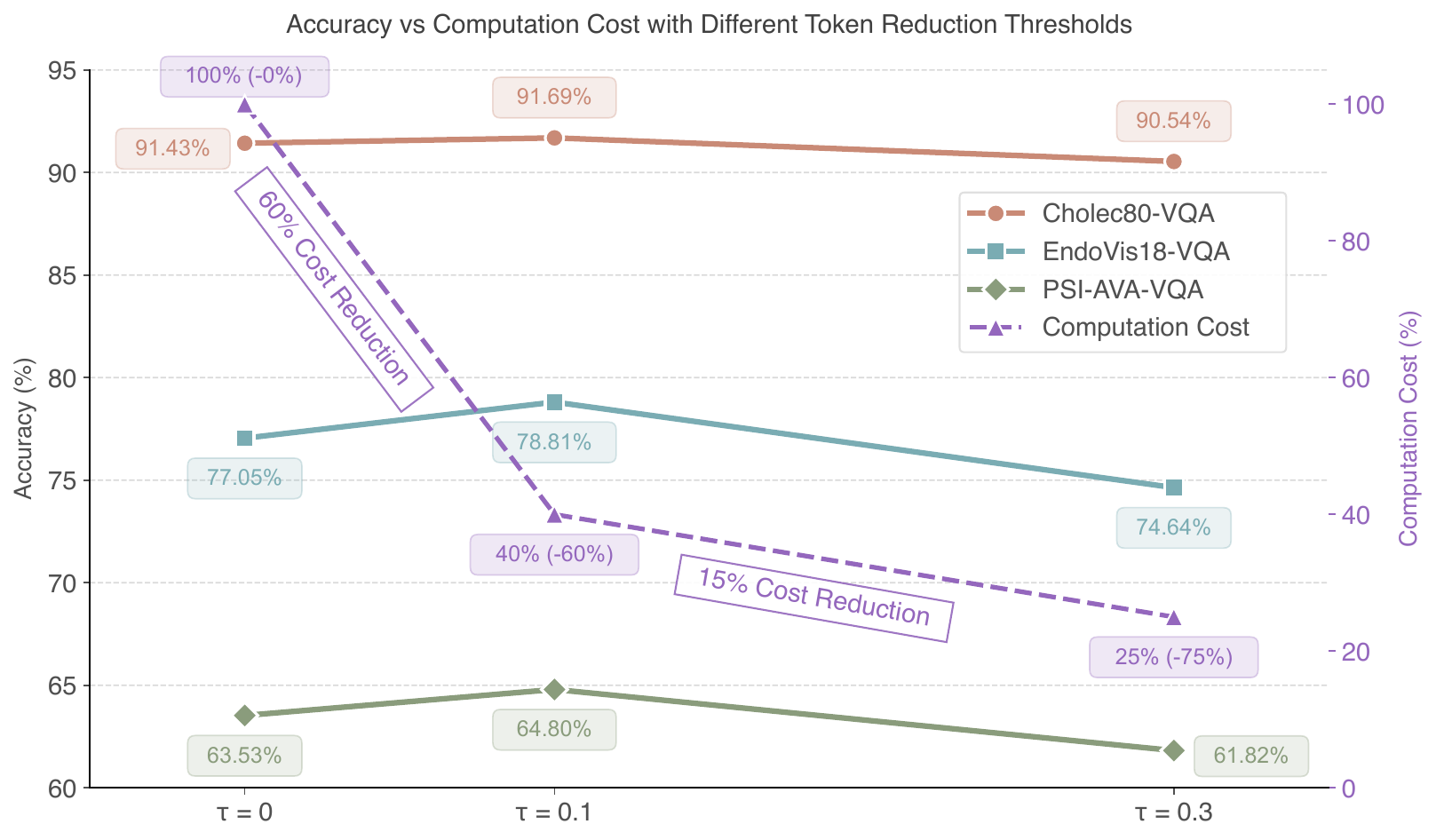}
\caption{Ablation study on token reduction threshold and cost comparison (Further results in Table~\ref{tab:token_red}).}
    \label{fig:ablation_thre}
\end{figure}

\section{Experiment}
\subsection{Settings}

We comprehensively evaluate OmniV-Med on seven medical benchmarks covering 2D/3D and video QA tasks. For 2D images, we follow LLaVA-Med and measure accuracy and recall on Rad-VQA~\citep{lau2018dataset}, SLAKE~\citep{liu2021slake}, and PathVQA~\citep{he2020pathvqa}. For 3D images, we evaluate on M3D-VQA tasks~\citep{bai2024m3d} with metrics consistent with previous studies. For videos, we report the accuracy on Cholec80-VQA~\citep{seenivasan2022surgical}, PSI-AVA-VQA~\citep{valderrama2022towards}, and EndoVis18-VQA~\citep{seenivasan2023surgicalgpt}.

We compare OmniV-Med against domain-specific SoTA baselines across medical 2D/3D and video tasks. For 2D images, we select Med-MoE~\citep{jiang2024med}, LLaVA-Med~\citep{li2024llava}, and BiomedGPT~\citep{zhang2023biomedgpt} as baselines. For 3D images, we compare with RadFM~\citep{wu2023towards} and M3D-LaMed~\citep{bai2024m3d}. In the video domain, we compare against Surgical-VQA~\citep{seenivasan2022surgical}, LV-GPT~\citep{seenivasan2023surgicalgpt} and Surgical-LLaVA~\citep{jin2024surgical}. 

\subsection{Main Results}
\begin{figure*}[!t]
    \centering
    \includegraphics[width=1\linewidth]{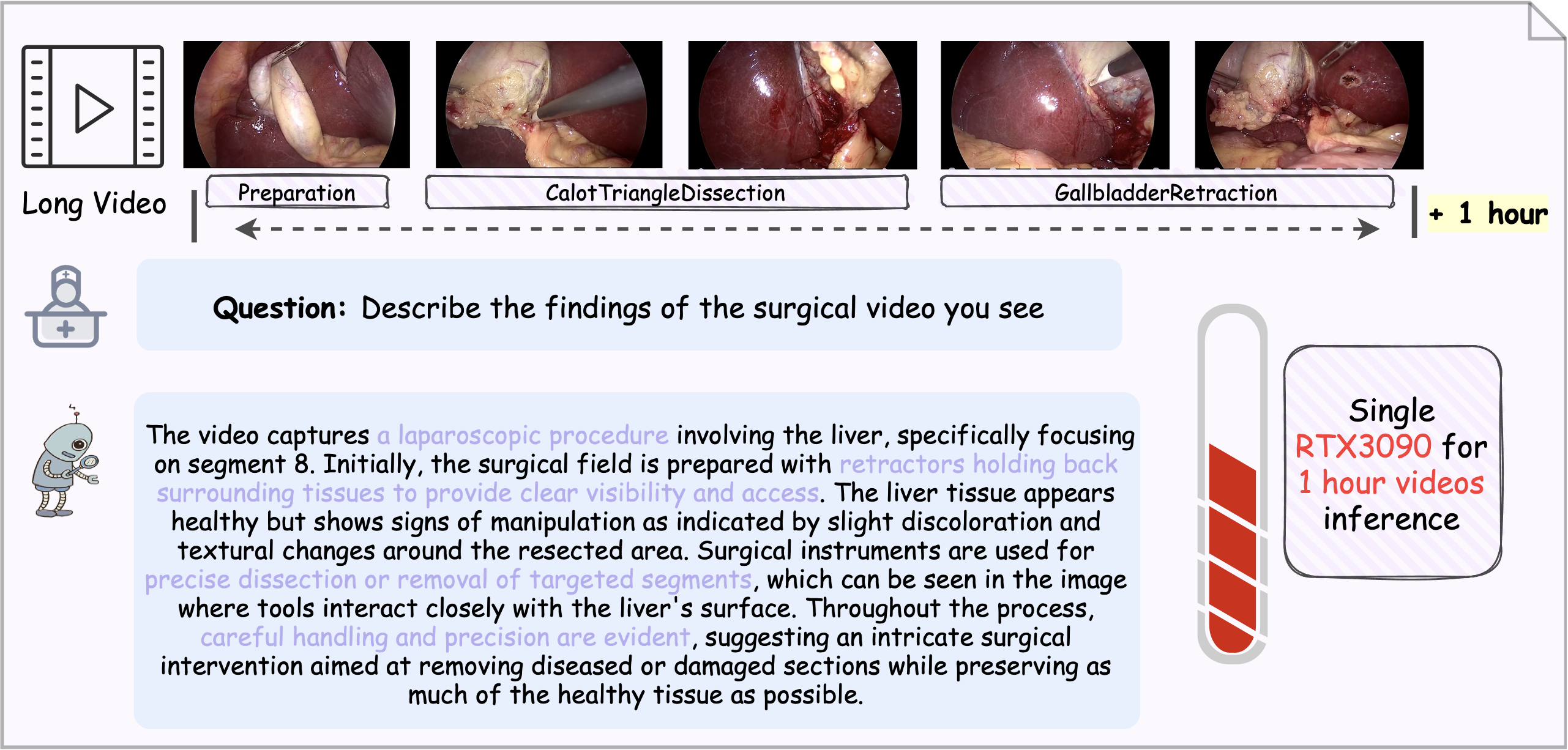}
\caption{OmniV-Med demonstrates robust performance on long-video inference with one single RTX 3090.}
    \label{fig:long_video}
\end{figure*}

\noindent \textbf{Consistent Improvements over all Modalities and Tasks.} Our experiments demonstrate that OmniV-Med achieves superior performance than modality-specific SoTA baselines across all benchmarks and modalities, highlighting the strong generalization capability. Additionally, our token pruning strategy significantly reduces the number of tokens by an average of 60\% for 3D images and medical videos, substantially lowering training and inference costs. This also enables support for long-video inference (see Figure~\ref{fig:long_video}), further demonstrate the universal capability of OmniV-Med. 

\subsection{Ablation Study and Analysis}
\noindent \textbf{Ablation of Arbitrary Resolutions.} To validate the effectiveness of 2D rotary position embeddings (RoPE) in learning relationships between image patches, we conducted experiments by replacing fixed position embeddings with RoPE while keeping the visual encoder unchanged. As shown in Table~\ref{tab:rope_advantage}, using the same data in Stage 1 and Stage 2, and training only on 2D images in Stage 3, we found that RoPE better captures relative positional relationships between patches, significantly improving performance on 2D tasks.

\noindent \textbf{Ablation of Medical-Aware Token Reduction.} We conduct experiments to evaluate the impact of token reduction on performance and computational cost. As shown in Figure~\ref{fig:ablation_thre}, the incorporation of token reduction significantly reduces visual tokens for 3D images and videos by an average of 60\%. More importantly, this pruning strategy, which leverages the characteristics of medical images, does not degrade performance. Instead, it helps the model focus on more critical regions of the images, leading to improved performance.

\noindent \textbf{Cost Analysis.} Training the OmniV-Med-1.5B model requires less than 24GB of GPU memory, due to our high-quality data, unified visual encoder, and medical-aware token reduction method. This enables training on eight RTX 3090 GPUs, while the 7B version supports inference on a single RTX 3090 GPU. Such efficiency enhances accessibility and underscores the practical value of our models for real-world clinical applications.

\begin{figure*}[!t]
    \centering
    \includegraphics[width=0.9\linewidth]{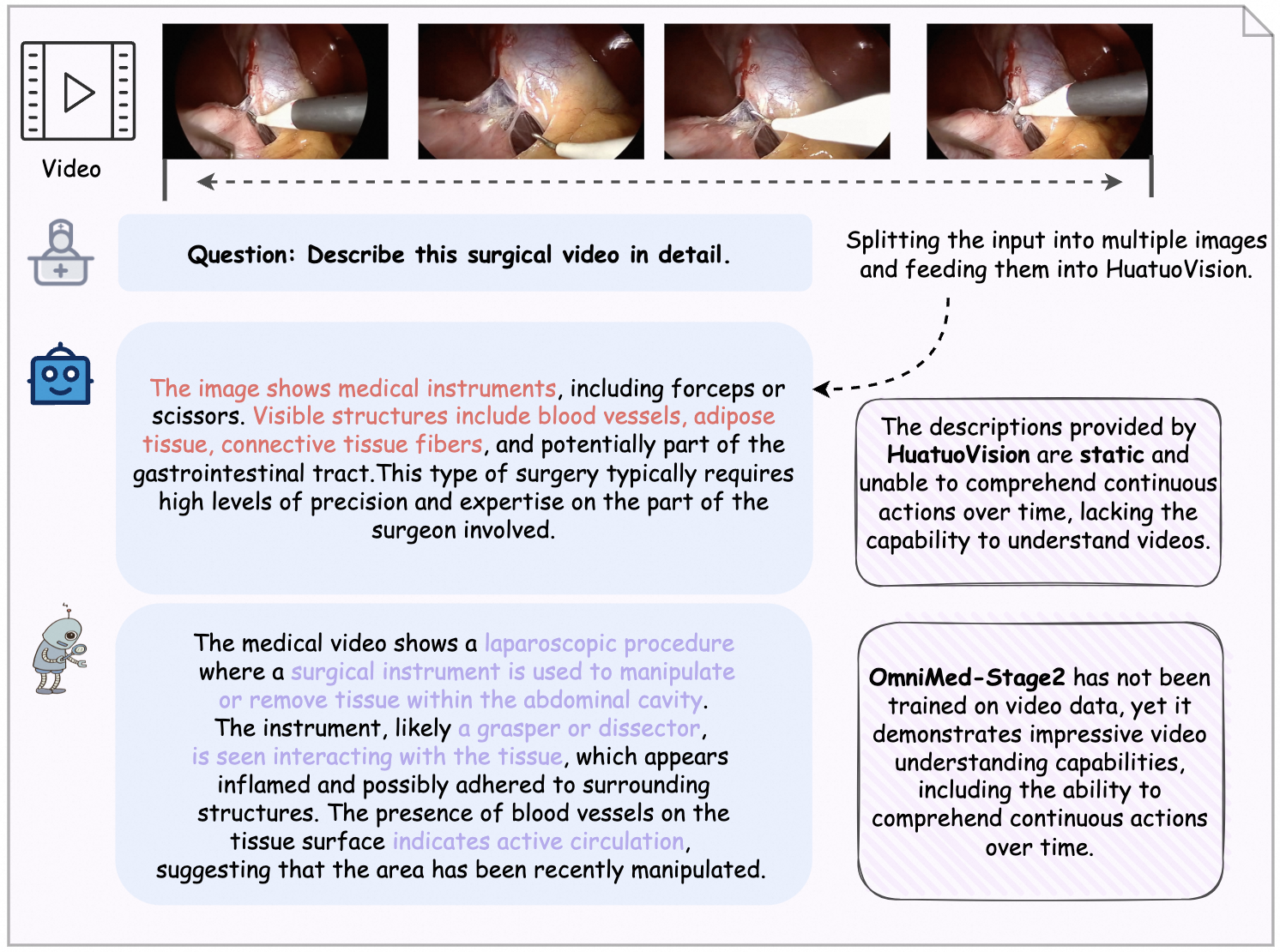}
\caption{OmniV-Med-Stage2 demonstrates outstanding task transfer capability, extending from 2D image understanding to video comprehension, showcasing a strong grasp of temporal action dynamics.}
    \label{fig:appendix_case1}
\end{figure*}

\noindent \textbf{Unified Visual Understanding Enables Task Transferring Capability.}
Our analysis reveals that by employing 2D RoPE, the model learns the relative positional relationships between patches in 2D medical images, which extends to understanding patches across multi-view (3D) and multi-time (video) dimensions. Specifically, even without video-specific training data during stage 1 and stage 2, OmniV-Med demonstrates exceptional comprehension of video content. As shown in Figure~\ref{fig:appendix_case1}, the model accurately interprets complex medical surgical procedures, highlighting significant improvements over traditional ViT-based approaches.

\begin{figure*}[!ht]
    \centering
    \small
    \includegraphics[width=0.9\linewidth]{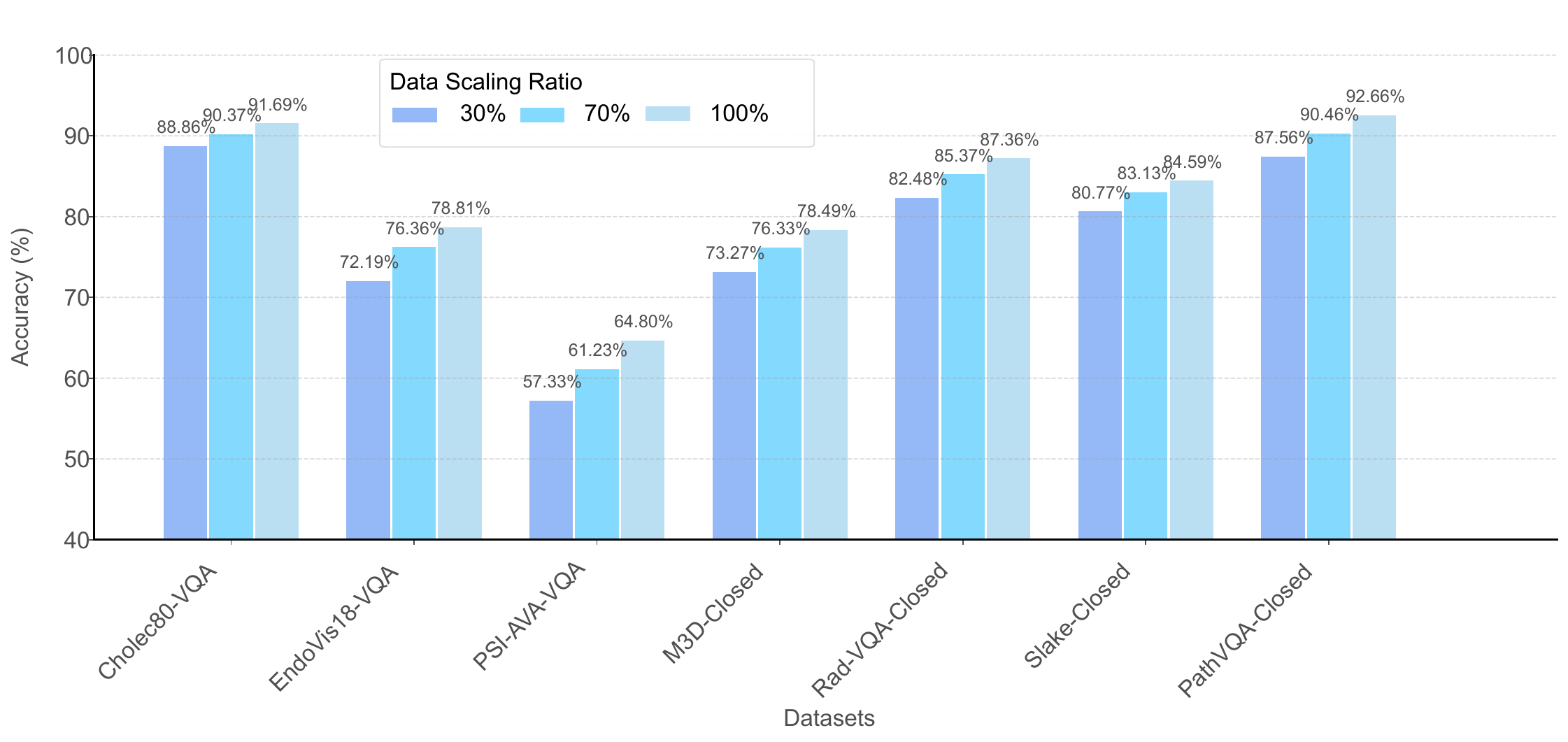}
\caption{Performance under different data scaling ratio.}
    \label{fig:Ablation_data_ratio}
    \includegraphics[width=0.9\linewidth]{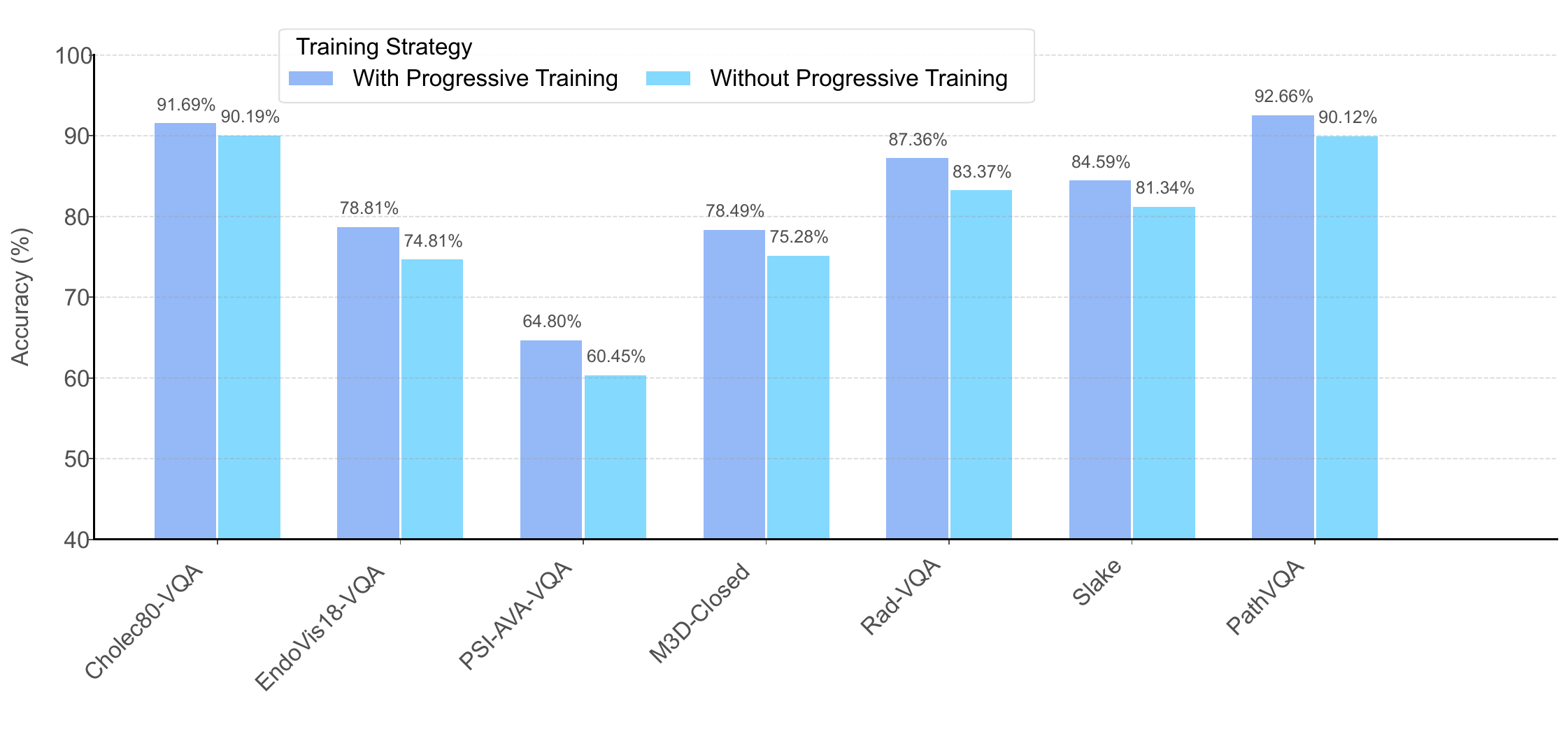}
\caption{Performance comparison with and without progressive training strategy.}
    \label{fig:Ablation_progress}
\end{figure*}

\noindent \textbf{Effect of Data Scaling.}
By incorporating diverse modalities in stage 3 on a unified visual encoder, we observe that introducing 3D images and videos alongside 2D images does not degrade performance on 2D benchmarks. Instead, it enhances performance, as shown in Figure~\ref{fig:Ablation_data_ratio}. This indicates that our training paradigm enables Med-VLMs to follow the data scaling law~\citep{kaplan2020scaling} in multimodal training scenarios, where increasing training data consistently improves performance. 

\noindent \textbf{Progressive Training Yields the Best Performance.}
We investigate the effect of jointly training stage 2 (2D images) and stage 3 (mixed-modal data) compared to our progressive training strategy. As shown in Table~\ref{fig:Ablation_progress}, the best performance is achieved when 2D image understanding is robustly established before extending to 3D images and videos. This underscores the importance of 2D image understanding as a foundation and validates the effectiveness of our progressive strategy.

\begin{figure}[!ht]
    \centering
    \includegraphics[width=1\linewidth]{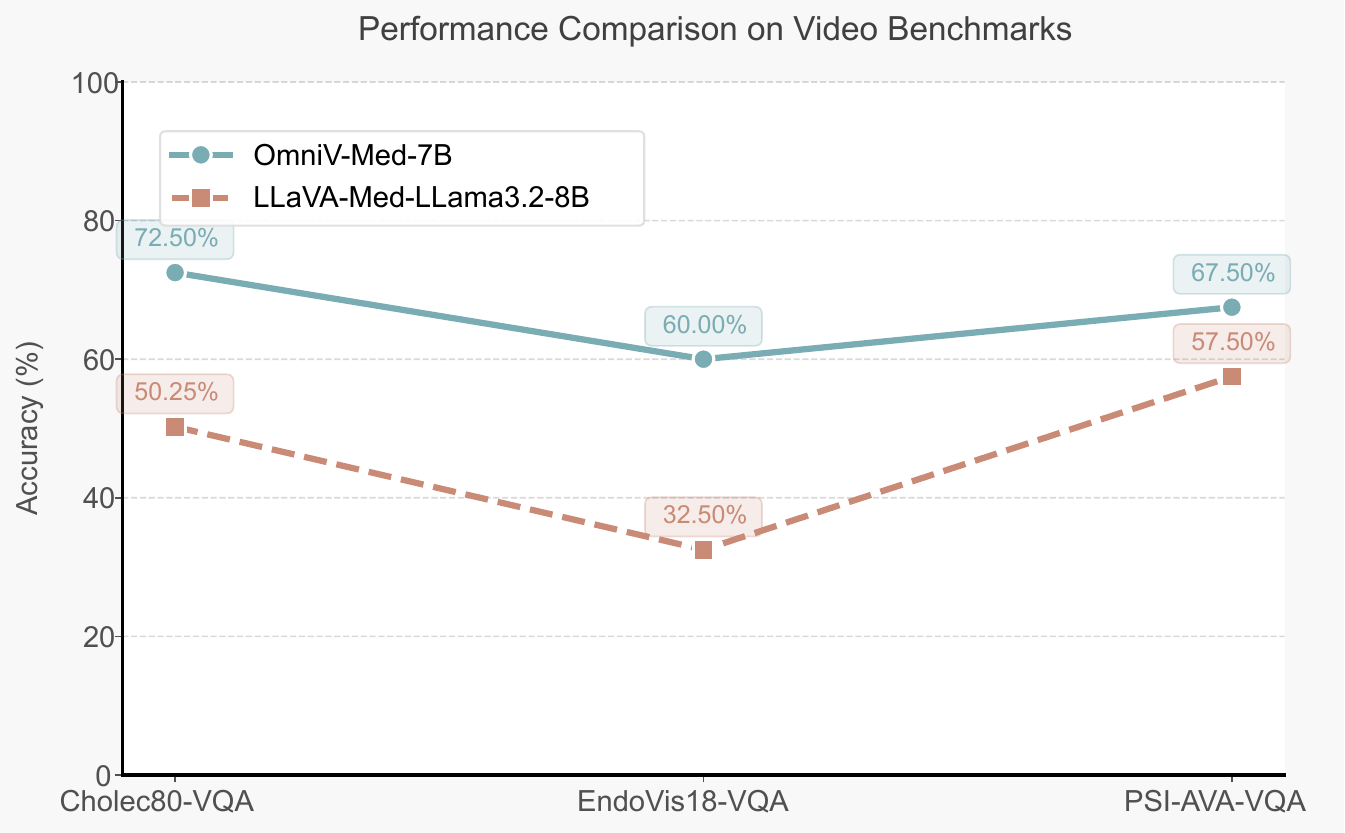}
\caption{Task transfering capablity of OmniV-Med.}
    \label{fig:task_trans}
\end{figure}
    
\noindent \textbf{Effect of Task Transferring Capability}
\label{details_stage2}
We conducted a zero-shot Video-VQA (Video Question Answering) capability comparison between OmniV-Med, trained only on 2D images through stages 1 and 2, and LLaVA-Med-LLaMA3.2~\citep{li2024llava}, which has demonstrated excellent performance in 2D images. We selected a total of 120 original VQA items from Cholec80-VQA~\citep{seenivasan2022surgical}, PSI-AVA-VQA~\citep{valderrama2022towards}, and EndoVis18-VQA~\citep{seenivasan2023surgicalgpt}, and reformulated them into multiple-choice questions where the model selects "yes" or "no" to evaluate its zero-shot VQA performance. As shown in Figure~\ref{fig:task_trans}, our OmniV-Med exhibited superior task transfer capability, with its rotary position-adaptive encoder demonstrating exceptional generalization ability.

\begin{table*}[htbp]
\centering
\begin{tabular}{ll|cc|cc|cc}  
\toprule
 & & \multicolumn{2}{c|}{VQA-RAD} & \multicolumn{2}{c|}{SLAKE} & \multicolumn{2}{c}{PathVQA} \\
\textbf{Embedding Type} & \textbf{Model} & \textbf{Recall} & \textbf{Accuracy} & \textbf{Recall} & \textbf{Accuracy} & \textbf{Recall} & \textbf{Accuracy} \\
\midrule
Fixed Positional & OmniV-Med-Tiny & 55.50 & 82.93 & 80.23 & 83.12 & 33.26 & 90.19 \\
Embedding &  & & & & & & \\
\hline
Rotary Position & OmniV-Med-Tiny & \textbf{59.22} & \textbf{84.93} & \textbf{84.59} & \textbf{87.98} & \textbf{38.87} & \textbf{92.66} \\
Embedding &  & \textbf{(+3.72)} & \textbf{(+2.00)} & \textbf{(+4.36)} & \textbf{(+4.86)} & \textbf{(+5.61)} & \textbf{(+2.47)} \\
\bottomrule
\end{tabular}
\caption{Ablation of Fixed Positional Embedding and Rotary Position Embedding.}
\label{tab:rope_advantage}
\end{table*}

\begin{table*}[htbp]
\centering
\begin{tabular}{l|ccc}
\toprule
\textbf{Threshold \& Method} & \textbf{Cholec80-VQA} & \textbf{EndoVis18-VQA} & \textbf{PSI-AVA-VQA} \\
\midrule
0           \quad OmniV-Med-Tiny                & 91.43                & 77.05                  & 63.53                \\
0.1 \quad OmniV-Med-Tiny     & \textbf{91.69} (\textbf{+0.26}) & \textbf{78.81} (\textbf{+1.76}) & \textbf{64.80} (\textbf{+1.27}) \\
0.3 \quad OmniV-Med-Tiny     & 90.54                & 74.64                  & 61.82                \\
\bottomrule
\end{tabular}
\caption{Performance Comparison on Video Datasets with Varying Token Reduction Thresholds.}
\label{tab:token_red}
\end{table*}

\subsection{Related Work}
\paragraph{Medical Large Vision-Language Models} Following the success of general-purpose Vision-Language Models (VLMs)~\citep{jiang2024modality,jiang2024joint,li2024llavaone,zhang2025videollama}, recent advancements in Medical Large Vision-Language Models have showcased specialized paradigms tailored for 2D, 3D, and video-based medical imaging~\citep{liu2024medcotmedicalchainthought,wu2023generalistfoundationmodelradiology}. For 2D imaging, models like LLaVA-Med~\citep{li2024llava} leverage multi-stage training (pretraining-modality alignment-fine-tuning) to align X-rays with diagnostic reports, while Med-MoE~\citep{jiang2024med} employs mixture-of-experts architectures to balance computational efficiency and diagnostic accuracy. In 3D imaging, M3D-LaMed~\citep{bai2024m3d} addresses volumetric complexity through 3D Vision Transformers and spatial pooling perceptrons, enabling cross-modal retrieval across CT/MRI datasets. RadFM~\citep{wu2023generalistfoundationmodelradiology} further pioneers multi-modal integration, combining natural language queries with 2D/3D radiologic images via adaptive fusion mechanisms. For surgical video analysis, frameworks like Surgical-LLaVA~\citep{jin2024surgicalllavasurgicalscenariounderstanding} are designed to handle surgical videos, providing real-time analysis and assistance during surgical procedures. Nevertheless, these methodologies possess inherent limitations: most models are restricted to unimodal visual analysis (e.g., X-ray or CT), lacking integration for multimodal diagnostic scenarios (e.g., 2D, 3D, Video). Single modality leads to inadequate cross-modal feature integration, while traditional architectures struggle to balance computational efficiency with anatomical detail preservation when handling high-resolution medical imaging~\citep{yang2024cross}.
\section{Conclusion}
In this paper, we construct the first multimodal instruction-following dataset containing 252K examples and train the OmniV-Med family of models, which includes two parameter sizes: 1.5B and 7B. These models are designed to flexibly support a wide range of practical clinical application scenarios. The OmniV-Med models achieve robust performance across diverse tasks involving 2D images, 3D volumes, and videos, demonstrating their versatility and effectiveness. Our work lays a solid foundation for the future development and practical deployment of Med-VLMs, paving the way for broader applications in clinical settings.

 \section*{Ethical Considerations}
In this study, we design a vision-language model with the potential for practical deployment in the medical domain, supporting multiple modalities of medical input, including 2D images, 3D volumes, and videos. Our work is grounded in the use of publicly available datasets that have been released by prior researchers, ensuring compliance with ethical standards. Specifically, all 2D and 3D images, as well as video data, are sourced from publicly accessible repositories and do not involve sensitive personal information or privacy concerns.

To further ensure ethical integrity, the dataset entries have been manually checked  to verify that they do not contain any content that could raise ethical or privacy issues. While our research leverages these well-established and pre-processed datasets, we recognize the importance of applying strict ethical considerations when extending our findings to other datasets, particularly those involving sensitive or private medical information. Robust safeguards must be implemented to protect patient confidentiality and prevent misuse of such data.

Additionally, as we advance the capabilities of vision-language models in the medical domain, it is crucial to balance these technological innovations with a careful assessment of their broader implications. This includes addressing potential challenges such as increased computational demands and their associated environmental impact, as well as ensuring equitable access to the benefits of these models across diverse healthcare settings.

\bibliography{custom}

\appendix



\end{document}